\documentclass[10pt]{article}
\pdfoutput=1

\usepackage{microtype}
\usepackage{graphicx}
\usepackage{subfigure}
\usepackage{booktabs}
\usepackage{makecell, relsize}
\usepackage{xspace}
\usepackage{multirow, enumitem}
\usepackage{tipa,changepage}

\usepackage{hyperref}
\usepackage{cleveref}
\usepackage{xcolor}
\usepackage[sort,comma,authoryear,round]{natbib}

\newcommand{\ra}{\rightarrow}
\newcommand{\png}[1]{figures/#1.png}

\renewcommand{\b}{\textbf}

\crefname{appsec}{Appendix}{Appendices}

\title{Low-Memory Neural Network Training: \\A Technical Report}

\author{%
  Nimit S. Sohoni, Christopher R. Aberger, Megan Leszczynski, \\Jian Zhang, Christopher R\'e \\
  \emph{Stanford University} \\
  \small nims@stanford.edu, caberger@alumni.stanford.edu, mleszczy@stanford.edu, \\
  \small zjian@stanford.edu, chrismre@cs.stanford.edu
}

\begin{document}

\maketitle

\begin{abstract}
Memory is increasingly often the bottleneck when training neural network models. 
Despite this, techniques to
lower the overall memory requirements of training have been less widely studied
compared to the extensive literature on reducing the memory requirements of inference.
In this paper we study a fundamental question: \emph{How much memory is }actually\emph{ needed to
train a neural network?} To answer this question, we profile the overall memory
usage of training on two representative deep learning benchmarks --- the WideResNet model
for image classification and the DynamicConv Transformer model for machine translation --- and
comprehensively evaluate four standard techniques for reducing the training
memory requirements: (1) imposing sparsity on the model, (2) using low precision,
(3) microbatching, and (4) gradient checkpointing. We explore how each of these
techniques in isolation affects both the peak memory usage of training
and the quality of the end model, and explore the memory, accuracy, and computation
tradeoffs incurred when combining these techniques.
Using appropriate combinations of these techniques,
we show that it is possible to the reduce the memory required to train
a WideResNet-28-2 on CIFAR-10 by up to 60.7x with a 0.4\% loss in accuracy, and reduce the
memory required to train a DynamicConv model on IWSLT'14 German to English translation
by up to 8.7x with a BLEU score drop of 0.15.
\end{abstract}

\section{Introduction}
\label{submission}

Recently, there has been an explosion of interest around techniques
to reduce the memory requirements of neural network \textit{inference} \citep{deepcompression,pruning}.
A central driver for this work is the fact that deep neural networks are notoriously parameter-hungry,
typically containing several millions or even billions of parameters \citep{gpt2},
making memory a significant hardware bottleneck for even storing and evaluating these models.
Unfortunately, \textit{training} is inherently more memory-intensive
than inference (often by orders of magnitude),
because it requires the storage of much more than just the network itself.
However, the majority of state-of-the-art techniques to reduce inference
memory are inapplicable to reducing training memory, as they often require
first training a full-size model that is later compressed. As a result, training
still requires over-provisioned and costly hardware resources
compared to the resources needed for inference \citep{nvidiaarch}.

Recent trends in deep learning suggest that this problem will continue to worsen.
Larger network architectures are becoming increasingly popular; these highly
overparameterized models have achieved great successes
in computer vision \citep{he2016deep,szegedy2017inception,han2016eie},
natural language understanding \citep{vaswani2017attention, devlin2018bert},
reinforcement learning \citep{silver2017mastering}, and more. However, the size of
neural networks is limited by the available memory on the device(s) used for
training. For certain datasets, such as video \citep{video} and high-resolution medical images
\citep{qiu2017learning}, the prohibitive memory requirements
of current training algorithms often mean that smaller, less expressive models must be
used or the inputs must be downsampled to a lower dimensionality, thus destroying
information \citep{jared, yi2017optimizing}.
As data collection systems continue to advance, the dimensionality
of real-world datasets continues to grow, leading to an ever more pressing need
for the ability to train larger and more complex models. This necessitates the development of
more memory-efficient training algorithms.

\begin{table*}
\setlength{\tabcolsep}{0.6em}
\setlength\extrarowheight{3pt}
\begin{center}
\begin{tabular}{ l l | c | c | c | c }
& & Sparsity & Low Precision & Microbatching & Checkpointing \\
\toprule
\multirow{3}{4em}{\textbf{Memory}}
& \emph{Model} & $\downarrow$ & $\downarrow$ & & \\ \cline{2-6}
& \emph{Optimizer} & $\downarrow$ & $\downarrow$ & & \\ \cline{2-6}
& \emph{Activations} & & $\downarrow$ & $\downarrow$ & $\downarrow$ \\ \cline{2-6}
\hline
\multicolumn{2}{c|}{\textbf{Computation}}
& $\downarrow$ & $\downarrow$ & & $\uparrow$ \\
\hline
\multicolumn{2}{c|}{\textbf{Accuracy}} & ? & ? & ? & \\
\bottomrule
\end{tabular}
\end{center}
\caption{The core components and techniques explored in our study. The main goal of our
study is to quantify the accuracy sacrificed while leveraging sparsity, low precision,
microbatching, and gradient checkpointing to reduce the overall memory used during training.
We explore each of the ``question marks" further in \Cref{sec:approach}, where we evaluate
how the corresponding techniques individually affect the memory usage of training
and the final accuracy. We also briefly discuss their effects on the computational cost of training.
In \Cref{sec:experiments}, we explore combining these techniques.}
\label{table:overall}
\end{table*}

In this paper, we study how best to address these issues by drastically reducing
the total memory required throughout the entire training process while still
preserving accuracy as much as possible. This result would have several benefits:
\begin{itemize}[noitemsep,topsep=0pt,parsep=0pt,partopsep=0pt,leftmargin=*]
  \item \textbf{Enabling new modeling breakthroughs.} On current,
  resource-rich hardware platforms, reducing
  the memory requirements of training would enable researchers to design
  and explore even larger and higher-capacity model architectures,
  on higher-dimensional datasets.
  \item \textbf{Training at the edge.} Edge computing devices are largely forsaken
  in favor of (resource-rich) data-center level hardware for training today.
  Reducing the memory requirements of training would facilitate training models from scratch
  (or fine-tuning existing models) on memory-poor devices, for instance smartphones.
  This would be advantageous to privacy (data can remain on-device)
  and to save on the costs of data transfer, and
  would also have positive impact in areas of interest
  such as federated learning \citep{konevcny2016federated}.
  \item \textbf{Paradigm shift for new hardware architectures.}
  Reducing the memory usage of training could help inspire the design of the next generation of hardware accelerators,
  as the monetary and energy cost, and potentially wall-clock time,
  of training could be reduced on appropriately designed hardware that take advantage of this
  \citep{prabhakar2017plasticine, masters2018revisiting}.
\end{itemize}

We focus on two standard benchmarks representative of different domains: training a WideResNet for image classification on
CIFAR-10,
and training a DynamicConv Transformer (DC-Transformer; often referred to as DynamicConv) \citep{payless}
language translation model on IWSLT'14 German to English (De $\ra$ En).
To set up our study of memory reduction techniques, we first classify the different types of memory that are used
during training. We categorize each component of memory as one of the following:
(1) \textbf{model memory}, which consists of the model parameters,
(2) \textbf{optimizer memory}, which consists of the gradients and momentum
vectors, and (3) \textbf{activation memory}, which consists of the intermediate network activations
(stored for reuse during backpropagation). The goal of our study is to
understand the accuracy tradeoffs when existing, well-studied
techniques to reduce these types of memory are used throughout training (see \Cref{table:overall}).
The key techniques we study are:

\begin{itemize}[noitemsep,topsep=0pt,parsep=0pt,partopsep=0pt,leftmargin=*]
\item \textbf{Sparsity (to reduce model and optimizer memory).} We use the
dynamic sparse reparameterization technique introduced by \citet{dsr} to
make the model parameters sparse \textit{throughout training},
which has the benefit of also making the gradients and momentum sparse. We find that
the WideResNet model can handle sparsity up to 70\% during training with an accuracy loss
of only 0.3\%, while the DC-Transformer can handle sparsity up to 60\% during training with a
BLEU score drop of 0.8.
Sparsity also reduces the overall number of floating point operations (FLOPs) required.
\item \textbf{Low precision (to reduce model, optimizer, and activation memory).}
Standard training algorithms use single precision (32 bits) for all quantities. However,
using half precision (16 bits) instead is a common technique for reducing memory requirements \citep{mixedprecision}.
We find that, despite the potential for numerical convergence issues, low precision does not significantly
affect the final accuracy, causing no drop in accuracy for WideResNet and an 0.15 drop in BLEU
score for DC-Transformer.
Low precision also reduces the amount of overall computation required.
\item \textbf{Microbatching (to reduce activation memory)}. In minibatch-based training algorithms,
one can either send the entire minibatch through the network at once and then update
the weights accordingly, or sequentially send smaller subsets (called microbatches) of the minibatch through the network,
accumulating the gradients until the entire minibatch has been processed \citep{huang2018gpipe}.
This is mathematically equivalent to standard training in models such as DC-Transformer which do not contain
batch normalization, and indeed causes no drop in accuracy for DC-Transformer. However,
microbatching changes the statistical properties of batch normalization layers; nevertheless, we find that
training with microbatch sizes as small as 10 does not lead to any loss of accuracy on WideResNet.
Microbatching trades a reduction in memory for a corresponding reduction in parallelism, but does not
change the overall amount of required computation (FLOPs).
\item \textbf{Gradient checkpointing (to reduce activation memory).}
Gradient checkpointing (or simply checkpointing) \citep{chen2016training, openai-blog} also reduces the amount of activation memory,
by only storing a subset of the network activations instead of all of the intermediate outputs (which is what is typically done).
As a consequence, the activations that are not stored must be recomputed during the backward pass.
Therefore, checkpointing increases the amount of computation; however,
checkpointing has no effect on the statistical accuracy as the training procedure is numerically unchanged.
Because there is no accuracy tradeoff, we instead present the computational tradeoff
introduced with this technique. We show that for a 30\% increase in FLOPs, checkpointing can reduce the memory
required for the activations by 5.8x on WideResNet and 5.7x on DC-Transformer.
\end{itemize}
These techniques were selected because they are simple, fundamental, and applicable to a wide variety
of network architectures. Despite this, there does not currently exist a comprehensive study of how
each technique affects memory and accuracy in different settings, or how best to combine these techniques.
In this paper, we systematically investigate each of these questions.

While many works have shown how to produce \textit{end models} with greatly reduced memory requirements, the training procedure used to create these models typically requires a much larger memory footprint. Few works have demonstrated an ability to directly train models under a strict \textit{total memory budget} while maintaining competitive accuracy. However, there has been partial progress toward this goal. The imposition of sparsity throughout training has been investigated by \citet{deepr, set, dsr, golub2019dropback, set2}. \citet{chen2016training} introduced checkpointing to reduce the memory required for the network activations. \citet{masters2018revisiting} investigated training with small batches. Crucially, however, none of these works address the reduction of \textit{all three} components of memory simultaneously. Training with low precision \citep{mixedprecision} can address all three components --- but only reduces the total memory requirements by a factor of at most 2x overall (when 16-bit precision is used for everything).

Reducing the total memory requirements of training by a significant amount is challenging because one must address all three components of memory; however, the various memory reduction techniques can interact in complex ways to affect the quality of the model, as we show with our experiments (\Cref{sec:experiments}). Therefore, naively picking the ``best" setting for each memory reduction technique in isolation can be suboptimal. We study the memory-accuracy \textit{tradeoffs} incurred by combining different techniques, in order to identify better heuristics for achieving the best possible accuracy using a fixed memory budget for training.
\\\\
\noindent Our contributions and an outline of the rest of the paper are as follows:

\begin{itemize}[noitemsep,topsep=0pt,parsep=0pt,partopsep=0pt,leftmargin=*]
  \item In \Cref{sec:background}, we describe and profile the major components of memory: model
  memory, optimizer memory, and activation memory. Although
  much effort to date has focused on reducing model memory,
  we show that while training both the WideResNet and DC-Transformer architectures, the model
  memory accounts for the smallest portion of overall memory usage, while the activations
  account for the largest.
  \item In \Cref{sec:approach}, we describe the four main techniques we study to
  reduce the memory usage of training:
  sparsity, low precision, microbatching, and gradient checkpointing.
  We evaluate each of these techniques in isolation and quantify
  the accuracy tradeoffs associated with using each
  technique throughout training. We also discuss how each technique
  affects the amount of computation required for training.
  \item In \Cref{sec:experiments}, we study how to best combine all four of the above techniques. We show that, when combined appropriately, these techniques can reduce the total memory requirements of training a WideResNet-28-2 on CIFAR-10 by up to 60.7x with an 0.4\% drop in accuracy. We compare sparsity to simply making the WideResNet smaller, and find that sparse WideResNets can achieve higher accuracy for the same memory budget. Next, we show that low precision, microbatching, and checkpointing can reduce the memory requirements of training a DC-Transformer on IWSLT'14 De $\ra$ En by up to 8.7x with a BLEU score drop of 0.15; however, we find that in this case, smaller DC-Transformer networks actually outperform the sparse ones.
\end{itemize}

\section{Preliminaries}
\label{sec:background}

To better understand the current memory requirements of training, we identify and profile the
main components of memory used while training neural networks. We first categorize
and describe the three main different types of memory that are used throughout
training. Next, we present memory profiles for two representative case studies:
training a WideResNet model for image classification and training a DC-Transformer for machine translation.
The results of this memory profiling inform the techniques and tradeoffs we study in \Cref{sec:approach}.

\subsection{Types of Memory}
The three main sources of memory consumption while training a neural network are as follows:

\paragraph*{Model Memory.}
The model memory is simply the memory used to store the model parameters, i.e. the weights and biases of each layer in the network.
This is the only major memory type that is common across both inference and training for the model architectures under study.

\paragraph*{Optimizer Memory.} Optimizer memory refers to the memory used to store the gradients and any momentum buffers during training. For instance, if standard SGD with momentum is used, there is one momentum value corresponding to each weight in the model. During backpropagation, we compute (a stochastic estimate of) the gradient with respect to all trainable model parameters. The momentum buffers are then updated based on this computed gradient.

\paragraph*{Activation Memory.} During training, the outputs, or \textit{activations}, of each layer in the network are stored for reuse in the backward pass of backpropagation (as they are required to compute gradients). This is what we refer to as the forward activation memory. (We also count the network inputs as part of the forward activation memory if they need to be stored for reuse in the backward pass.) During backpropagation, we will also compute gradients with respect to these layer outputs; we refer to this as the backward activation memory. The sum of these is the activation memory.

\subsection{Profiling}
\label{sec:profiling}
To better understand how to reduce the memory requirements of training, we first profile how much each of these three types of memory contributes to the total memory consumption. The results of our profiling are shown in Figure \ref{fig:pie}.

\subsubsection{Types of Memory}
\begin{figure*}
\vskip 0.2in
\begin{center}
\centerline{\includegraphics[scale=0.4]{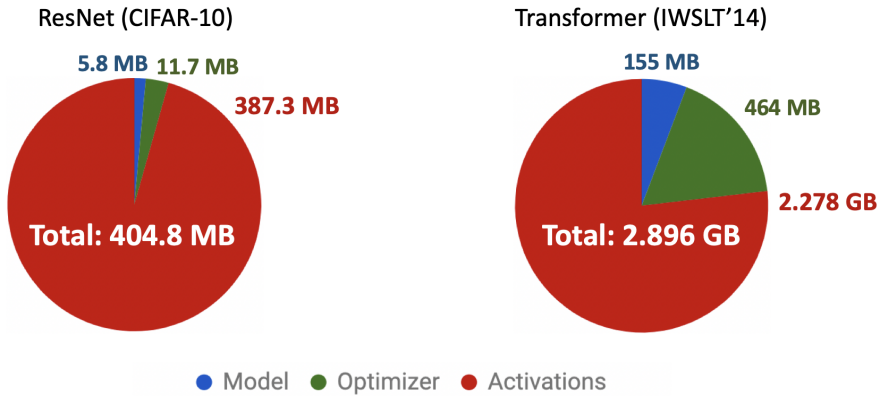} }
\end{center}
\vskip -0.2in
\caption{Pie charts of training memory consumption for (a) WideResNet on CIFAR-10 [left] and (b) DC-Transformer on IWSLT'14 German to English [right].}
\label{fig:pie}
\end{figure*}

\paragraph*{Model Memory.} Figure \ref{fig:pie} shows that the model memory accounts
for the smallest amount of memory of the three classes of memory
we have described. Although the model memory is 2-3x less than the optimizer memory
for our settings and can be orders of magnitude less than the activation memory, it is the
memory type that is most commonly optimized for because it is the component that is
common across both inference and training.

\paragraph*{Optimizer Memory.} In our study, the optimizer memory is always 2 or 3x more than
the model memory, depending on the optimizer used. On the WideResNet architecture,
it is 2x the memory for the model parameters because we run SGD with Nesterov momentum. For this
optimizer, we need one buffer to store the gradients and an additional buffer to store the
momentum values.\footnote{It is also possible to only store the momentum buffer without
a separate gradient buffer, and update this buffer in place during backpropagation. However,
this means that standard gradient clipping cannot be done.
It also has the potential to lead to numerical issues, especially if low precision is used. We do not
study this approach in our work.}
On the DC-Transformer model architecture, we use the Adam
optimizer \citep{kingma2014adam}, which requires one buffer for the gradients
and two momentum buffers (first and second moment estimates), resulting in a memory usage
that is 3x that of the model.

\paragraph*{Activation Memory.}
As is apparent from Figure \ref{fig:pie}, the activations consume most of the memory in standard training of these architectures. However, this need not always be the case: it is partly due to the fact that batch sizes as large as possible are typically used for training to maximize throughput on GPUs. In fact, as we show in \Cref{sec:microbatching}, the number of activations can be reduced significantly without loss of accuracy.

Unless the model architecture has a very large amount of branching, the backward activation memory is in general much smaller than the forward activation memory; the gradients with respect to the forward activations have the same sizes as the corresponding forward activations themselves (by definition), but are only computed during the backward pass and do not need to be stored after they have been used to backpropagate through a layer.

\paragraph*{Other.}
There is some additional fundamental memory overhead not included in these three categories
(for instance, temporary storage buffers required for single computations).
Nevertheless, these additional intermediates comprise a small fraction of the total memory usage,
while the three components of memory described above account for the vast majority.
Therefore, we direct our efforts to reducing these components, noting that in many cases similar approaches
 can be used to reduce the impact of the additional overheads as well.

\subsubsection{Profiler Implementation}
\label{sec:profiler}
Model and optimizer memory can readily be calculated by summing up the memory required for each model and optimizer tensor (which can be calculated as \texttt{tensor.numel() * tensor.element\_size()} in PyTorch).

To calculate the activation memory, we traverse the autograd graph generated by PyTorch and sum up
the necessary storage space for the activation inputs of all operations. (Several operations - among others,
transpose operations and addition operations - do not require any input activations to be stored. Some others,
such as ReLU, only require 1 bit per input element \citep{gist}, and so on. We hardcode in these exceptions.)
More details are provided in \Cref{app:profiler-impl}.

We compare the calculated memory to the output of the PyTorch memory profiling tool
\texttt{torch.cuda.max\_memory\_allocated}, which computes the maximum memory for
all currently allocated tensors at any time, and find that the total of the model, optimizer, and activation
memory we calculate matches the PyTorch profiling results within 5\% on WideResNet and 11\% on DC-Transformer.

\section{Techniques and Tradeoffs}
\label{sec:approach}

\begin{figure*}
\vskip -0.2in
\begin{center}
\centerline{\includegraphics[scale=0.3]{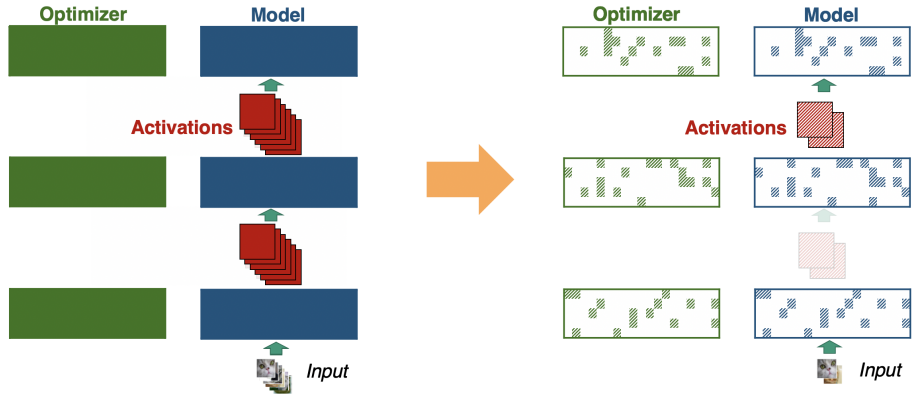}}
\caption{Schematic describing approach. The left-hand side represents standard training: the model and the optimizer state are dense, batch sizes are large, and single precision is used. The right-hand side represents our integrated approach: the model and optimizer state are sparse (rectangles are mostly white instead of completely filled), the batch size is small (fewer activations), and all quantities are in low precision (diagonal fill instead of full shading). Checkpointing is used, so the lightened activations on the right are not stored and instead recomputed on-the-fly.}
\label{fig:schematic}
\end{center}
\vskip -0.2in
\end{figure*}

We study the fundamental tradeoffs associated with using the following four
memory reduction techniques during training:
(1) sparsity, (2) low precision, (3) microbatching,
and (4) gradient checkpointing. First, we briefly present each technique and describe the type(s) of memory it reduces.
Next, we assess the tradeoffs associated with each individual technique,
for a WideResNet model on the CIFAR-10 image classification task.
For each technique besides checkpointing, there
is no adverse tradeoff with computation
(in terms of the total number of floating point operations --- see \Cref{table:overall}),
so we primarily study their memory-accuracy tradeoffs.
Conversely, for checkpointing, there is no memory-accuracy tradeoff,
so we present the memory-computation tradeoff that it incurs.
The tradeoffs presented here form the basis for our
end-to-end experimental results in \Cref{sec:experiments} that combine all techniques.
Figure \ref{fig:schematic} illustrates the four techniques.

\subsection{Sparsity}
\label{sec:sparsity}
A \textit{sparse} tensor is a tensor in which most of the entries are zero. The zero entries do not need to be explicitly stored, thus reducing the memory usage. It is well known how to apply sparsity to compress \textit{trained} neural networks with little or no loss of accuracy \citep{pruning}. However, techniques that maintain sparsity \textit{throughout the entire training process} are more recent.

\paragraph*{Technique.} The sparse training technique we leverage in this paper is dynamic sparse reparameterization (DSR), introduced by \citet{dsr}, which is a modification of SET \citep{set}. This technique initializes the network with a fixed sparsity pattern, and after every few iterations (based on a preset schedule), the smallest-magnitude weights of the network are pruned (set to zero) and an equal number of new nonzeros are introduced throughout the network to replace them. Parameters are pruned based on a global threshold, which is scaled at each rewiring step to attempt to prune a fixed preset number of parameters each time. This prune-and-replace operation is known as \textit{rewiring}. The frequency at which to rewire and the number of parameters to rewire (specified as a fraction of the overall network size) are important hyperparameters of this procedure. By design, the total number of nonzeros in the model is constant throughout training. Gradients are only computed with respect to the nonzero parameters, which means that the gradients are also sparse.

In our reimplementation, we make a small modification to the algorithm of \citet{dsr}: we reset the momentum buffers to their initial value (typically zero) after each rewiring. Thus, the momentum buffers are also sparse in addition to the model and gradient --- in fact, the model, gradient, and momentum all share the same sparsity pattern --- so DSR reduces both the model and optimizer memory.

\begin{figure}[ht]
\begin{center}
\centerline{\includegraphics[scale=0.5]{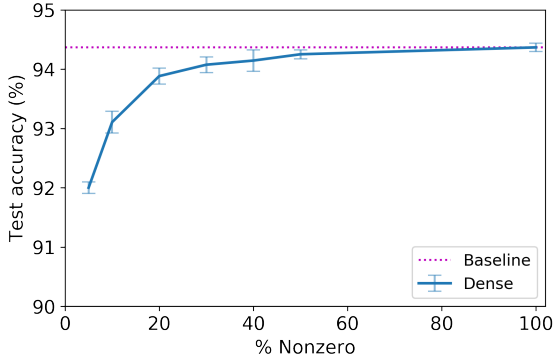} }
\caption{Test accuracy vs. percentage of nonzero parameters, for WRN-28-2 on CIFAR-10. Vertical lines denote 95\% confidence intervals. Dotted line is the dense baseline accuracy.}
\label{fig:sparsity}
\end{center}
\vskip -0.2in
\end{figure}

\paragraph*{Accuracy Tradeoff.} To evaluate how sparsity affects model accuracy, we fix the same base WideResNet model architecture and vary the allowed percentage of nonzeros in convolutional layers\footnote{As in \citep{dsr}, we do not sparsify the first convolutional layer, the classification layer, batch normalization layers, or biases. Over 99.6\% of the parameters of the dense model are in the tensors we do sparsify.} from 100\% (dense baseline) all the way down to 5\%. For each level of sparsity, we search for the best initial learning rate in \{0.025, 0.5, 0.1, 0.2\}.\footnote{The default initial learning rate from \citep{dsr} is 0.1.} Then, we grid search both the \textit{frequency} at which to rewire in \{0.125, 0.25, 0.5, 1, 2, 4\} times the default frequency from \citep{dsr} and \textit{fraction} of parameters to rewire in \{0.25, 0.5, 1, 2, 4, 8\} times the default fraction from \citep{dsr} (see \Cref{app:hyperparams} for more details). All grid searches are done on the validation set; we then train a model on the full training dataset with the selected hyperparameters, and evaluate it on the test set (see \Cref{app:validation} for more details). The results are presented in Figure \ref{fig:sparsity}. Consistent with \citep{dsr}, we observe that sparsification down to 20\% nonzero entries (80\% sparsity) in convolutions causes less than a 0.5\% drop in test accuracy over the dense baseline; below this point, the accuracy begins to degrade more rapidly.

\paragraph*{Computation.} Using sparse operations reduces the number of FLOPs,
as the zero entries can be ignored.
However, in general, sparsity leads to irregular memory access patterns, so
it may not actually result in a speedup; the performance depends heavily on
the level of sparsity and the hardware being used.

While sparse matrix multiply kernels exist for both CPU and GPU (e.g.\ cuSPARSE \citep{cusparse}),
hardware support for sparse convolution is currently limited. Nevertheless, some recent works have
shown that using \textit{direct sparse convolution}, modest speedups can be
achieved over dense convolution even on current-generation CPU and GPU hardware
\citep{directsparseconv, escort}. These direct approaches have the additional
advantage of bypassing the memory-costly \texttt{im2col} operation. This
exciting work suggests that on future or special-purpose hardware
architectures, the computational performance gains from sparsity could be
even greater.

\paragraph*{Storage.} For each sparse tensor, both the actual nonzero entries of the tensor, as well as the indices corresponding to these nonzero entries, need to be stored. Fortunately, as the gradient and momentum have the same sparsity pattern (i.e. set of nonzero locations) as the model, they can all share the same index arrays.

In the direct sparse convolution algorithms cited above,
a 4-dimensional convolutional weight tensor
of shape $c_o \times c_i \times k_1 \times k_2$, where
$c_o,c_i$ are respectively the number of output and input channels and
$k_1, k_2$ are the kernel sizes, is stored as a sparse matrix of shape
$c_o \times (c_i k_1k_2)$. In other words, the $i^{th}$ row of the matrix is the
flattened version of the 3-D tensor corresponding to the $i^{th}$ output channel.
This matrix is stored in compressed sparse row (CSR) format.
Note that we only need $\lceil{\log_2(c_ik_1k_2)}\rceil$ bits to store each column
index entry. We adopt this convention to calculate the storage size
of our sparse weight tensors.

\subsection{Low Precision}
\label{sec:precision}

Training neural networks is typically done in \textit{single precision}
(FP32) [IEEE 32-bit] arithmetic. However, recent work \citep{mixedprecision} suggests that
\textit{half precision} (FP16) [IEEE 16-bit] arithmetic is generally sufficient for training.
Using lower-precision arithmetic both reduces all components of memory
and decreases the total amount of required computation;
however, in some cases it can cause numerical issues \citep{mixedprecision}.

\paragraph*{Technique.} Basic FP16 training is as simple as specifying the \texttt{float16} datatype for all weights, optimizer buffers, and inputs, and is supported by modern deep learning frameworks such as TensorFlow \citep{tensorflow2015-whitepaper} and PyTorch \citep{pytorch}. In addition, as suggested in \citep{apex}, we use dynamic loss scaling to avoid numerical underflow and overflow in gradients, and keep all batch normalization parameters in FP32. However, unlike in \citep{apex}, we do not keep a separate FP32 copy of weights, in order to conserve memory.

\paragraph{Accuracy Tradeoff.} We find that training in half precision does not meaningfully affect the final accuracy. We fix all hyperparameters except the precision and compare the accuracy achieved when training with single-precision (FP32) to that achieved using half-precision (FP16). When using FP32, the WideResNet model attains a test accuracy of $\mathbf{94.37 \pm 0.07\%}$ on CIFAR-10, and when using FP16, the model attains a test accuracy of $\mathbf{94.43 \pm 0.17\%}$.

\paragraph*{Computation.} While using half-precision arithmetic does not reduce the number of FLOPs, as the same number of floating-point operations are performed, it does reduce the cost of \textit{each} floating-point operation on hardware that natively supports it \citep{nvidiavolta}. The exact amount of speedup is dependent on the hardware used.

\subsection{Microbatching}
\label{sec:microbatching}
As shown in Figure \ref{fig:pie}, the memory required for computing and storing the network activations can far outstrip the memory required to store the model itself. Two possible solutions are to downsample the input data to a smaller size, or to make the model ``narrower" (for instance, reducing the number of output channels of each convolution). Beyond a point, each of these techniques necessarily leads to a loss of accuracy --- the former because it discards information in the data, and the latter because it reduces the expressivity of the model.

As the number of activations is directly proportional to the minibatch size, the minibatch size can instead be reduced to reduce the memory usage. However, this changes the optimization properties of the training procedure, potentially necessitating significant hyperparameter tuning. Alternatively, \textit{microbatching} \citep{huang2018gpipe} can be used: the minibatch is split into smaller groups called microbatches, which are each run independently forward and back through the network. The gradients are accumulated in a separate buffer until the desired number of examples is processed, and only then are the parameters and momentum buffers updated (and the gradient buffer zeroed). If the examples within a batch are independent, this is mathematically (but not necessarily numerically) equivalent to the original procedure.

\begin{figure}[ht]
\vskip 0.2in
\begin{center}
\centerline{\includegraphics[scale=0.5]{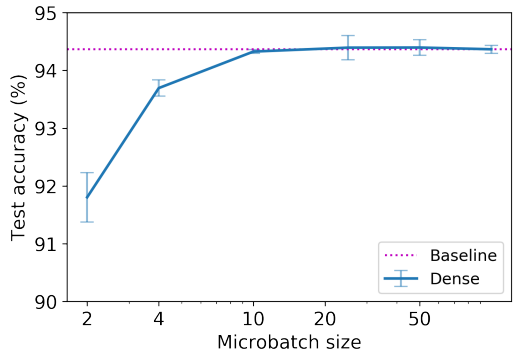}}
\caption{Test accuracy vs. microbatch size, for WRN-28-2 on CIFAR-10. Vertical lines denote 95\% confidence intervals. Dotted line is the baseline with a single microbatch, i.e. same minibatch and microbatch size of 100.}
\label{fig:batchsize}
\end{center}
\vskip -0.2in
\end{figure}

However, if the model contains batch normalization layers \citep{batchnorm} -- which WideResNet does -- the examples within a minibatch are no longer independent. Batch normalization is ubiquitous in state-of-the-art models in many domains, especially image classification; it normalizes the activations at each layer across all the examples in each minibatch. Thus, the microbatch normalization statistics will differ from the statistics of the entire minibatch, which can affect the final accuracy of the model. Indeed, the smaller the microbatch, the less accurate the normalization statistics will be as estimates of the population statistics \citep{batchrenorm}. We investigate the effect of microbatching on model accuracy in \Cref{fig:batchsize}.

\paragraph{Technique.}
For the experiments in this section, we simulate microbatching by splitting each minibatch into smaller groups over which normalization statistics are calculated, but still running the entire minibatch forward and backward through the network at the same time rather than sequentially processing each microbatch (the latter is ``true microbatching"). In other words, all the microbatches are run through the network together, but normalized independently as they are in true microbatching. The simulation approach therefore takes as much memory as standard training; it was done to save on computational time for our experiments, since GPUs are highly optimized for minibatch parallelism rather than frequent sequential computation. It is mathematically equivalent to true microbatching.

However, this simulation technique is not necessarily \textit{numerically} equivalent to true microbatching.
The reason for the numerical difference is that in standard training,
the accumulation of gradients across samples in a minibatch
is often done implicitly via the matrix multiply (GEMM) operation. This technique is
numerically beneficial because it leverages the fused-multiply-add (FMA)
operation, which can typically utilize a ``wide" accumulator
(for instance, a 32-bit accumulator as in the Volta GPU architecture \citep{nvidiavolta}),
thus reducing the overall floating point error from this reduction operation.
This is also true of our simulation approach.
By contrast, when the microbatches are run sequentially instead of in parallel,
this accumulation is instead effectively done elementwise across
each microbatch in the minibatch. This means our effective accumulator width
when using FP16 with microbatching is only 16 bits.
Nevertheless, in \Cref{sec:resnet-micro2}, we conduct experiments in FP16 comparing our simulated microbatching
technique to true microbatching, and find that differences in accuracy between the two methods are within at most 0.06\%.

\paragraph{Accuracy Tradeoff.}
In Figure \ref{fig:batchsize}, we plot the test accuracy versus the microbatch size used during training for WideResNet on CIFAR-10
(which contains batch normalization). We fix all other hyperparameters and vary the microbatch size.
We observe that models trained with microbatch sizes as small as 10 exhibit no loss in accuracy compared to the baseline
microbatch size of 100, although the accuracy begins to degrade significantly as the microbatch size is reduced even further.
So, although smaller microbatches have more highly variable normalization statistics as described earlier, this does not
have a deleterious effect on accuracy until the microbatch size is very small (4 or 2).

\paragraph*{Computation.} Microbatching does not affect the total amount of computation (FLOPs) done during training;
rather, it reduces memory usage at the cost of reducing parallelism.
GPUs are optimized to be very efficient at regular, parallel computations; therefore, standard practice is often to use
as large a batch size as possible. However, alternative accelerators such as the recently proposed IPU \citep{masters2018revisiting}
are designed to be fast and efficient even with small batch sizes; thus, microbatching is well-suited
for such devices. In addition, for certain models and datasets, even state-of-the-art GPUs may not be able to fit large batches
in memory; microbatching allows for training in such memory-constrained settings.

\subsection{Checkpointing}
\label{sec:checkpointing}
\textit{Gradient checkpointing} \citep{chen2016training, openai-blog} is an algorithm to save memory on the network activations by only storing the activations of a subset of layers, rather than those of each layer as usual. The activations that are discarded are recomputed when necessary during backpropagation. Checkpointing is particularly convenient because the outcome of the forward and backward passes is mathematically \textit{and} numerically equivalent whether or not checkpointing is used, and so there is no memory-accuracy tradeoff. However, it is not a panacea, as, depending on the dataset, model architecture, and minibatch size, even the activations of a \textit{single layer} can take up a large amount of memory. This is often the case when dealing with, for instance, high-resolution or three-dimensional medical imaging data \citep{segmentation, blumberg2018}.

\paragraph{Technique.}
For example, a simple checkpointing strategy is to store the input activations of every $m$ nodes in the network during the forward pass, where a ``node" is defined as any operator that requires the inputs to be stored in order to compute its gradients. We denote this strategy \textsc{checkpoint-every-$m$}.
The nodes for which inputs are stored are called ``checkpoint nodes." In the simplest case of a feedforward network with no branching and $m \times n$ nodes, denoted $N_1, \dots, N_{mn}$, we might only store the inputs to $N_{m}, N_{2m}, \dots, N_{m(n-1)}$. During the backward pass, we recompute and store the inputs of all $m$ nodes between $N_{m(n-1)+1}$ and the output layer $N_{mn}$ in order to compute gradients with respect to these nodes, starting from $N_{m(n-1)}$. After this, we discard the inputs of these layers and then recompute the inputs of all nodes between $N_{m(n-2)+1}$ through $N_{m(n-1)-1}$, and so on. This means we need to store $n+m$ activation quantities at a time rather than $mn$, at the cost of an additional amount of recomputation up to the amount of computation in the standard forward pass.

Another strategy, \textsc{checkpoint-no-bn}, is to store all activations except the outputs of normalization operations (such as batch normalization) and subsequent ReLU operations, as these are computationally inexpensive to recompute \citep{inplacebn} compared to convolutional and fully-connected layers. This reduces the number of stored activations by nearly 50\% on WideResNet with a very small increase in FLOPs (less than 1\%). A more aggressive checkpointing strategy (suggested by \citep{openai-blog}) is to store only the outputs of each residual block, recomputing quantities in between consecutive checkpoint nodes (i.e. inside the residual branch) as necessary. More generally, we can store the output of every $m$ residual blocks for even more memory savings; we refer to this strategy as \textsc{checkpoint-residual-$m$} (store the output of every $m$ residual blocks). Note that all these techniques can be applied to the DC-Transformer model as well, as it also has residual connections in each encoder and decoder layer.

Finally, an even more sophisticated strategy is recursive: store the output of every $m$ residual blocks, only store the other residual block outputs in between checkpoints when recomputing \citep{openai-blog}, and always discard and recompute batch normalization outputs. We apply this technique to WideResNet; when backpropagating through a residual block, we keep the output of its first convolution in memory, but recompute through batch normalization layers on-the-fly as before. We refer to this strategy as \textsc{checkpoint-residual-$m$}*.

The default strategy, \textsc{checkpoint-none}, is to simply store all the activations that are necessary for reuse during backpropagation. This is done by default in most deep learning frameworks, and is the baseline we compare to. 

\begin{figure*}
\vskip 0.2in
\begin{center}
\centerline{\includegraphics[scale=0.5]{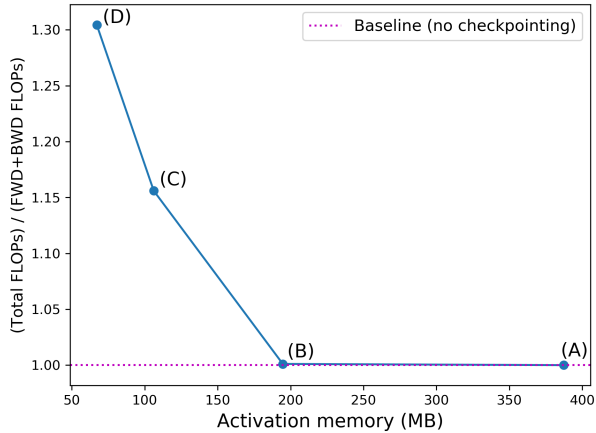}}
\end{center}
\vskip -0.2in
\caption{FLOPs ratio to baseline for various degrees of checkpointing, for training WideResNet on CIFAR-10. The total number of FLOPs is the number of FLOPs for the regular forward and backward pass, plus the number of FLOPs needed to recompute unstored values. The baseline is no checkpointing, which means that no recomputation is done. Memory values are based on 32-bit precision and a microbatch size of 100. On the plot, (A) denotes \textsc{checkpoint-none}, (B) denotes \textsc{checkpoint-no-bn}, (C) denotes \textsc{checkpoint-residual-1*}, and (D) denotes \textsc{checkpoint-residual-2*}.}
\label{fig:ckpt}
\end{figure*}

\paragraph{Computation Tradeoff.}
As described in the previous paragraph, checkpointing increases the amount of computation done, as some quantities computed during the forward pass must be recomputed. In \Cref{fig:ckpt}, we plot the FLOPs ratio to the baseline approach versus the peak memory usage for the network activations for the following checkpointing strategies (from most to least memory): \textsc{checkpoint-none}, \textsc{checkpoint-no-bn}, \textsc{checkpoint-residual-1*}, and \textsc{checkpoint-residual-2*}. We observe that, for instance, using \textsc{checkpoint-residual-2*} we can reduce the storage space for the activations by 5.8x with only a 30\% increase in total FLOPs.\footnote{Here, the FLOPs we consider are the total FLOPs for the forward pass, backward pass, and any necessary forward recomputations. We do not consider the FLOPs involved in taking the actual optimization step, i.e. updating the momentum buffers and the weights given the model gradient; these are the same regardless of the checkpointing strategy used.}

\section{Experiments}
\label{sec:experiments}
In \Cref{sec:cifar}, we evaluate the four techniques enumerated above on the WideResNet architecture for image classification on the CIFAR-10 task as before; in \Cref{sec:iwslt}, we do the same for the DynamicConv Transformer architecture for machine translation on the IWSLT'14 German to English translation task. For both tasks, we measure the memory and accuracy resulting from training with different settings of the techniques, and compare against the baseline of simply making the original network smaller (i.e. fewer or lower-dimensional layers).

\subsection{Experimental Setup}
We use the PyTorch framework \citep{pytorch}, version 1.0.1, and NVIDIA\textsuperscript{\textcopyright} V100 GPUs \citep{nvidiavolta}, with CUDA Toolkit 10.1 and cuDNN v7.4, for all experiments.

\subsubsection{Implementation Details}
We calculate the memory usage for the baseline training procedure using the method discussed in \Cref{sec:profiler}.

\noindent Below, we describe how we implement each of the four techniques in our experiments.

\paragraph{Sparsity.} As support for sparse tensor operations is currently limited, we simulate sparsity by representing each sparse tensor using a dense tensor of equivalent size and a binary mask indicating the nonzero locations (as done in \citep{stateofsparsity, dsr}). However, as our focus is on the \textit{fundamental} memory requirements of training, we compute the required model and optimizer memory usage by calculating the storage space for sparse tensors as if they were indeed stored in the flattened CSR format described in \Cref{sec:sparsity}.

\paragraph{Low precision.} In PyTorch, training with 16-bit precision (FP16) simply involves calling \texttt{.half()} on all model parameters, optimizer tensors, and input data.

\paragraph{Microbatching.} Unless explicitly noted, for the experiments in this section we simulate microbatching (for any given microbatch size), as per \Cref{sec:microbatching}.

\paragraph{Checkpointing.} As checkpointing does not affect the final accuracy and increases the computation time, we do not actually integrate it into our training code; rather, we separately calculate the total activation memory and FLOPs required for different checkpointing strategies using our profiler described in \Cref{sec:profiler}.
\\\\
We compute the total (peak) memory usage numbers reported below by summing the model, optimizer, and activation memory calculated as described above.

\subsubsection{Datasets}
\paragraph*{Image Classification.} We run experiments on the CIFAR-10 dataset \citep{cifar}, which consists of 50,000 training and 10,000 test datapoints. The inputs are 32 x 32 RGB images, and the label is one of 10 classes. We use standard data augmentations of per-channel image normalization, horizontal flipping, and random crops of size 32 x 32 after padding by 4 pixels on each side, as in \citep{dsr}.
We compute the test accuracy after each epoch and report the best value, as is common practice in the image classification setting \citep{dsr, imagenet-in-minutes}.

\paragraph*{Neural Machine Translation.} For translation, we run experiments on the IWSLT'14 German to English translation task \citep{iwslt}.
We replicate the setup of \citep{payless}. The training set consists of 160K sentence pairs. All data is lowercased, and the vocabulary is a 10K joint source and target byte pair encoding (BPE). To evaluate, we follow standard practice \citep{payless, ott2018scaling} and average the last 10 model checkpoint files, and use this averaged model to compute the BLEU score on the test set.

\subsection{Image Classification Results}
\label{sec:cifar}
In this section, we show how to combine all four techniques while training a WideResNet architecture on the CIFAR-10 image classification task. We show how to reduce the memory requirements of training by over 60x and achieve 93.95\% classification accuracy, competitive with the baseline accuracy of 94.37\%. We compare these techniques to the simple alternative of making the base network architecture smaller, and find that they generally outperform this alternative, i.e. attain higher accuracy for a fixed memory budget.

\subsubsection{Setup}
\paragraph{Model.}
Our base model architecture is a WideResNet-28-2 (WRN-28-2) \citep{wideresnet} with identity shortcuts, containing approximately 1.46M parameters in total. This model is the same as the one used in \citep{dsr} except that all shortcuts are identity shortcuts in our model, which reduces the number of parameters by about 11K. We chose this model to ensure a strong baseline, as it is already quite parameter-efficient yet still attains good accuracy; thus, further compression of this model is meaningful.

\paragraph{Training.} As in \citep{dsr}, we use SGD with Nesterov momentum \citep{nesterov} as our optimizer. Default hyperparameters are the same as in \citep{dsr}; a list of hyperparameters is given in \Cref{app:hyperparams}.

\subsubsection{Tradeoff Studies}
In this section, we study the memory-accuracy tradeoffs of combining \textit{multiple} memory reduction techniques during training. We show that by appropriately combining all four techniques we have discussed, we can reduce the combined model, optimizer, and activation memory requirements to just 6.7 MB (60.7x reduced compared to the original 404.8 MB) and achieve 93.95\% classification accuracy (compared to the original 94.37\%).

\begin{figure}[ht]
\vskip 0.2in
\begin{center}
\centerline{\includegraphics[scale=0.5]{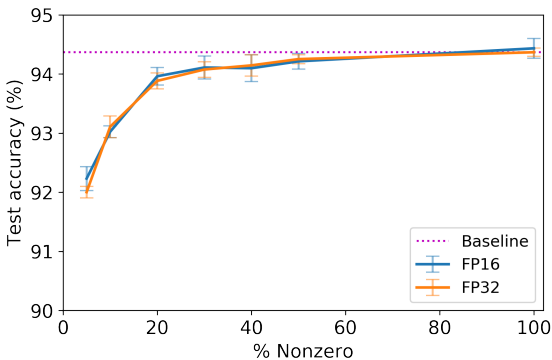} }
\caption{Plot of test accuracy vs. sparsity for WRN-28-2 on CIFAR-10, for FP16 and FP32. The lines align extremely closely with one another, meaning that half precision does not significantly affect the accuracy.}
\label{fig:resnet-sparsity-lp}
\end{center}
\vskip -0.2in
\end{figure}

In \Cref{fig:resnet-sparsity-lp}, we compare training with different levels of sparsity in both FP16 and FP32. For each level of sparsity, we use the learning rate, rewiring frequency, and rewiring fraction selected via grid search in \Cref{sec:sparsity} (see \Cref{app:validation} for more details). We find that there is no significant difference between FP16 and FP32 in terms of accuracy, as suggested by \Cref{sec:precision}.\footnote{Although the presented results are on the test set, we first compared FP16 and FP32 on the validation set and also observed no significant difference in accuracy.} Therefore, we use FP16 for all subsequent experiments in this section.

We now study the effects on the accuracy and total memory usage as we vary the sparsity level and microbatch size. We use the checkpointing strategy \textsc{checkpoint-residual-2*}, which reduces the activation memory by approximately 5.8x and increases FLOPs by approximately 30\%. Given that training with FP16 operations is cheaper than training in FP32, the overall amount of computation still compares favorably to that of the standard training approach.

\begin{figure}[ht]
\vskip 0.2in
\begin{center}
\centerline{\includegraphics[scale=0.5]{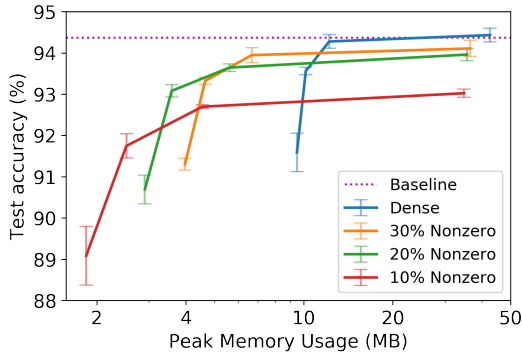} }
\caption{Plot of test accuracy vs. training memory usage for WRN-28-2 on CIFAR-10. Each curve represents a different sparsity level, while different points on a given curve represent different microbatch sizes (100, 10, 4, and 2). All settings use FP16 and \textsc{checkpoint-residual-2*}.}
\label{fig:resnet}
\end{center}
\vskip -0.2in
\end{figure}

In Figure \ref{fig:resnet}, we plot the accuracy versus the total memory budget of training for various sparsity levels and microbatch sizes. Again, for each level of sparsity we use the learning rate, rewiring frequency, and rewiring fraction selected in \Cref{sec:sparsity}.

From the figure, it is evident that the appropriate \textit{combination} of techniques is necessary to get memory-accuracy Pareto-optimal results (i.e. the best accuracy for a given memory budget, or the least memory required to attain given accuracy); simply increasing the sparsity level and keeping the microbatch size fixed, or vice versa, generally results in worse accuracy for a given memory budget than combining both techniques. A selection of Pareto-optimal datapoints from Figure \ref{fig:resnet} are also listed in Table \ref{table:pareto}. For instance, using a 30\% nonzero (70\% sparse) network, a microbatch size of 10, and \textsc{checkpoint-residual-2*}, training can be done in just over 6.7 MB of memory --- a \b{60.7x} reduction in memory usage compared to the 404.8 MB baseline --- with only an 0.4\% loss in accuracy compared to the baseline accuracy of 94.37\%. Compared to the baseline, the activation memory is reduced by 115x, the model memory by 4x, and the optimizer memory by 6x.

\subsubsection{Microbatching with Low Precision}
\label{sec:resnet-micro2}
To validate our microbatching simulation approach (where microbatches are normalized independently, but run through the network in parallel), we also studied the potential numerical issues caused by microbatching and low precision. As discussed in \Cref{sec:microbatching}, in our simulation approach, the effective accumulator width for the gradient calculations is 32 bits, versus 16 bits for true microbatching. We train with each of the two approaches to check whether this actually makes a difference in the final accuracy.

For $n = 4, 10$, we ran ``true microbatching" experiments with FP16, a microbatch size of $n$, and the standard minibatch size of 100, in which the examples were actually run sequentially and independently through the network (which means that $100/n$ gradients are accumulated before taking the gradient step). For comparison, we ran separate experiments with the same settings, except with our simulated microbatching approach. The results are shown in \Cref{table:micro}. We observe that for a microbatch size of 10, the difference between the true microbatching approach and our simulation approach is only 0.06\%, within the range of random variation; for a microbatch size of 4, the true microbatching approach actually outperforms the simulation approach by 0.06\%, also within the range of random variation. Thus, our simulated approach appears to be well predictive of the results with true microbatching.

\aboverulesep=0ex
\belowrulesep=0ex
\renewcommand{\arraystretch}{1.12}
\begin{table}[h]
\setlength{\tabcolsep}{5pt}
\begin{center}
\begin{tabular}{| c | c | c |}
\toprule
Microbatch Size & Test Acc. - True Microbatching & Test Acc. - Simulated Microbatching \\
\midrule
4 & $93.62 \pm 0.19$ & $93.56 \pm 0.09$ \\
10 & $94.21 \pm 0.12$ & $94.28 \pm 0.17$ \\
\bottomrule
\end{tabular}
\caption{Comparison between true microbatching (running microbatches sequentially) and simulated microbatching (running all microbatches together but normalizing separately), for CIFAR-10 WideResNet training. All settings use FP16 and a dense (100\% nonzero) network.}
\label{table:micro}
\end{center}
\vskip -0.3in
\end{table}
\aboverulesep=0.4ex
\belowrulesep=0.65ex
\renewcommand{\arraystretch}{1.0}

\subsubsection{Baselines}
\begin{figure}[ht]
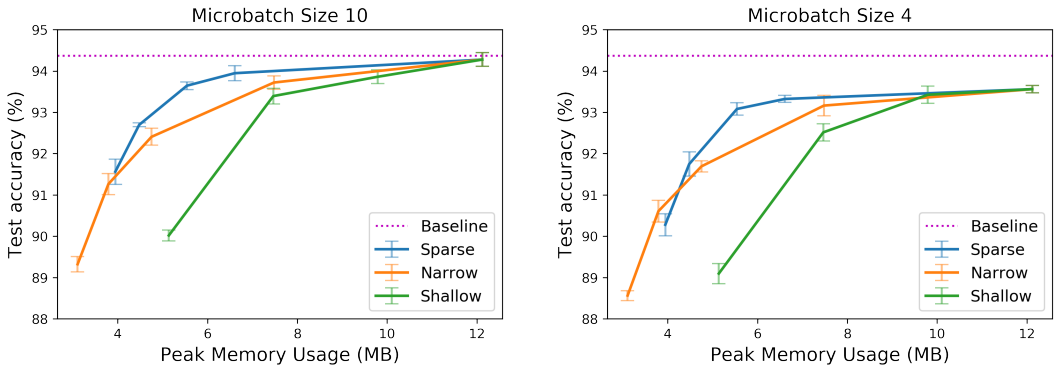

\vskip 0.2in
\begin{center}
\centerline{\includegraphics[scale=0.5]{\png{mb10}} \quad \includegraphics[scale=0.5]{\png{mb4}}}
\caption{Plot of test accuracy vs. training memory usage for WideResNet on CIFAR-10, when using a microbatch size
of 10 (left) or 4 (right). The three curves in each plot are generated by varying the number of layers (``shallow"),
the width of the layers (``narrow"), or the sparsity level of the layers (``sparse").
All settings use FP16 and \textsc{checkpoint-residual-2*}.}
\label{fig:resnet-compare}
\end{center}
\vskip -0.2in
\end{figure}

In Figure \ref{fig:resnet-compare}, we compare sparsity to the alternative of simply reducing the size of the network architecture. The most straightforward ways to reduce the network size are to reduce the depth (number of layers), or to reduce the width (i.e. the number of channels in convolutional layers). For each different network architecture, we search for the best initial learning rate in \{0.025, 0.5, 0.1, 0.2\}; as done for sparsity levels, these searches are done on the validation set, using full precision training without microbatching, and the optimal settings identified for each architecture are then used for all experiments on the test set (see \Cref{app:validation} for more details).

We observe that, up to a point, for a given memory budget, sparsity preserves the accuracy better than reducing either the network depth or width. However, reducing the network depth or width also reduces the number of activations, so, beyond a point, small enough network architectures necessarily outperform larger sparse model architectures for a fixed total training memory budget due to the discrepancy in activation storage. Nevertheless, as illustrated by the figure, this crossover happens when the memory budget is very low ($< 5$ MB), at which point the accuracy has already degraded significantly. We note that sparsity and network architecture modifications such as reducing the size or number of layers are orthogonal techniques; it is also possible to apply both concurrently, although we do not investigate this in this work.

\aboverulesep=0ex
\belowrulesep=0ex
\renewcommand{\arraystretch}{1.12}
\begin{table}[h]
\setlength{\tabcolsep}{5pt}
\begin{center}
\begin{tabular}{| c | c | c | c | c | c | c |}
\toprule
Nonzero \% & Microbatch Size & Memory & Mem. Reduction & Test Acc. \\
\midrule
100 & 100 & 404.8 MB & - & $94.37 \pm 0.07$ \\
\hline
100 & 100 & 42.6 MB & 9.5x & $94.43 \pm 0.17$ \\
100 & 10 & 12.2 MB & 33.2x & $94.28 \pm 0.17$ \\
30 & 10 & 6.7 MB & 60.7x & $93.95 \pm 0.18$  \\
20 & 10 & 5.6 MB & 72.2x & $93.64 \pm 0.09$ \\
20 & 4 & 3.6 MB & 113.0x & $93.08 \pm 0.15$ \\
10 & 4 & 2.5 MB & 160.8x  & $91.75 \pm 0.29$ \\
\bottomrule
\end{tabular}
\caption{Pareto-optimal settings for CIFAR-10 WideResNet training. The baseline, which is dense (100\% nonzero) and uses 32-bit precision, a minibatch size of 100, no microbatching, and no checkpointing, is the top row. Memory reduction is computed relative to this baseline. All other settings use 16-bit precision, and \textsc{checkpoint-residual-2*}. Accuracies are reported as percentages; accuracies exceeding the standard-length training baseline are bolded. (Note: Nonzero \% refers to the percentage of nonzeros in convolutions.)}
\label{table:pareto}
\end{center}
\vskip -0.3in
\end{table}
\aboverulesep=0.4ex
\belowrulesep=0.65ex
\renewcommand{\arraystretch}{1.0}

\subsubsection{Takeaways}
We find that the appropriate combination of sparsity, low precision, microbatching, and checkpointing can lead to a significant reduction in the overall (peak) memory usage of training. As shown in \Cref{table:pareto}, low precision and checkpointing alone yield a memory reduction of 9.5x with no accuracy loss compared to the baseline. Incorporating microbatching can boost this to 33.2x with a very minor accuracy loss of 0.09\%, and using sparsity can boost this to 60.7x with 0.42\% accuracy loss. In conclusion, we show that the current memory requirements of training can be significantly reduced on this dataset with little loss of accuracy.

\subsection{Machine Translation Results}
\label{sec:iwslt}
In this section, we evaluate the same set of techniques for training the DynamicConv Transformer (DC-Transformer) architecture \citep{payless} on the IWSLT'14 German to English (De $\ra$ En) translation task. We show that, using low precision, microbatching, and checkpointing we can reduce the memory requirements of training by 8.7x and achieve a BLEU score of 34.99, compared to the baseline BLEU score of 35.15. We again compare sparsity to making the base network architecture smaller. In this case, we find that the latter can be a better option than sparsity; however, low precision, microbatching, and checkpointing are all still highly advantageous.

\subsubsection{Setup}
\paragraph{Model.}
Our base model architecture is the DC-Transformer architecture from \citep{payless}. This architecture is similar to the original Transformer architecture proposed in \citep{attention}, but with dynamic convolutions \citep{payless} instead of self-attention layers; this modification was shown by \citet{payless} to achieve near-state-of-the-art results on the IWSLT'14 De $\ra$ En task, with fewer parameters than the original Transformer architecture. The baseline DC-Transformer model contains approximately 38.7M parameters. We use the \texttt{fairseq} library \citep{ott2018scaling} as a basis for our implementation.

\paragraph{Training.}
As in \citep{payless}, we use the Adam optimizer \citep{kingma2014adam}. Hyperparameters not under evaluation are kept the same as in \citep{payless}; more details are provided in \Cref{app:hyperparams}. To evaluate, we average the last 10 model checkpoint files, as done in \citep{payless, ott2018scaling}, and compute the BLEU score on the held-out test set.

\paragraph{Additional Methodology Details.}
For our sparsity experiments, we sparsify the weights of all fully connected layers and embeddings. The layers we sparsify contain over 98\% of the total parameters in the network.\footnote{Note that as we average the last 10 model checkpoint files before evaluation, the final model will be \textit{dense}. Nevertheless, the overall required memory usage of inference on this dense model is still less than that of training, as there is no optimizer state to store and activations do not need to be stored for backpropagation.}

When training in FP16, in addition to the procedure described in \Cref{sec:precision}, we up-cast the gradients, weights, and momentum to FP32 to compute momentum and weight updates before casting the weights back to FP16, as done by default in the \texttt{fairseq} library. As this can be done on a per-tensor basis or even on-the-fly, it can be done with insignificant memory overhead. In addition, we first scale up the momentum tensors to properly fit into the FP16 range before casting back to FP16. (This is not done in the \texttt{fairseq} library, which always stores the momentum in FP32.) We do this because we found that when simply rounding the momentum buffers back to FP16 without scaling, a large proportion of values were small enough that they were rounded to zero, causing training to quickly diverge.

\subsubsection{Tradeoff Studies}
\begin{figure}[ht]
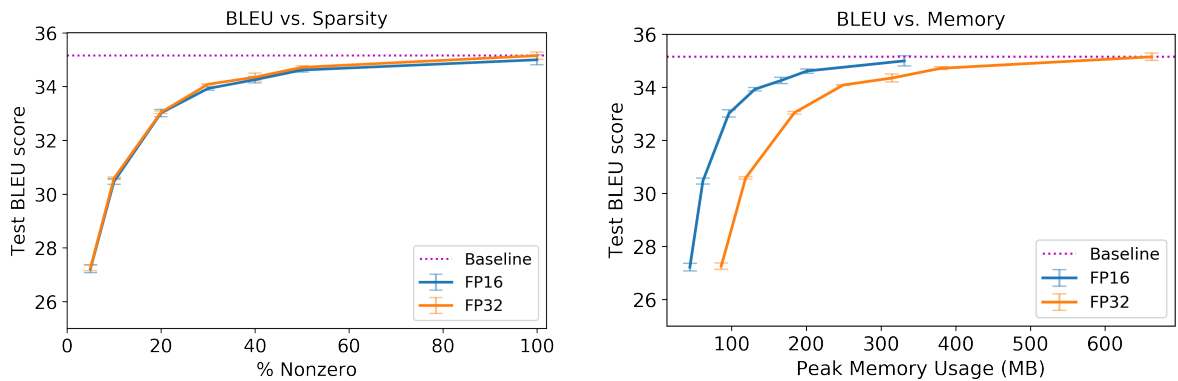

\vskip 0.2in
\begin{center}
\centerline{\includegraphics[scale=0.5]{\png{transformer_sparsity}} \quad \includegraphics[scale=0.52]{\png{transformer_memory}} }
\caption{(a) Plot of test BLEU score vs. \% nonzero for DC-Transformer on the IWSLT'14 German to English translation task. (b) Plot of test BLEU score vs. training memory usage [using the same experimental settings as those in (a), but plotted versus memory usage instead of sparsity]. For plot (b), memory numbers correspond to a microbatch size of 250 tokens and the checkpointing strategy \textsc{checkpoint-residual-1}.}
\label{fig:transformer}
\end{center}
\vskip -0.2in
\end{figure}

We evaluate the memory-accuracy tradeoffs of combining sparsity and low precision. For each sparsity level, we search the base learning rate in \{0.5, 1, 2, 4, 8\} times the default. We then grid search both the \textit{frequency} at which to rewire (in \{$<$no rewiring$>$, 0.125, 0.25, 0.5, 1, 2\} times the default frequency from \citep{dsr}) and \textit{fraction} of parameters to rewire (in \{0.25, 0.5, 1, 2, 4, 8\} times the default fraction from \citep{dsr}). All tuning is done on the validation set as described in \Cref{app:validation}; we then use the selected learning rate and rewiring hyperparameters for both FP32 and FP16 experiments on the test set. The results are displayed in Figure \ref{fig:transformer}(a) and Table \ref{table:transformer}. We observe that, as with WideResNet, training in FP16 instead of FP32 causes very little drop in the quality of the model as measured by BLEU score. However, the effects of sparsity on the model quality are more pronounced than on the WideResNet model.

In the DC-Transformer model, there is no batch normalization; instead, layer normalization is used, which does not introduce any interdependences between minibatch examples. Therefore, microbatching is mathematically (but not necessarily numerically) equivalent to standard training. In the translation setting, the smallest possible microbatch size is the number of tokens of the longest sentence in the dataset, as each sentence is considered a separate example. To adhere to this requirement, we therefore use a microbatch size of 250 tokens. We fix this as the microbatch size for all experiments in this section. We also use the checkpointing strategy \textsc{checkpoint-residual-1}, which results in approximately a 5.7x decrease in activation memory at the cost of approximately a 30\% increase in total FLOPs.

\aboverulesep=0ex
\belowrulesep=0ex
\renewcommand{\arraystretch}{1.12}
\begin{table}[h]
\setlength{\tabcolsep}{5pt}
\begin{center}
\begin{tabular}{| c | c | c | c | c | c |}
\toprule
Nonzero \% & Precision & Memory (MB) & Mem. Reduction & Test BLEU \\
\midrule
100 & 32 & 662 MB & 4.4x & $35.15 \pm 0.14$ \\
100 & 16 & 331 MB & 8.7x & $34.99 \pm 0.19$ \\
50 & 32 & 380 MB & 7.6x & $34.71 \pm 0.06$ \\
50 & 16 & 201 MB & 14.4x & $34.61 \pm 0.08$ \\
40 & 32 & 315 MB & 9.2x & $34.35 \pm 0.15$ \\
40 & 16 & 166 MB & 17.4x & $34.25 \pm 0.12$ \\
30 & 32 & 249 MB & 11.6x & $34.08 \pm 0.17$ \\
30 & 16 & 131 MB & 22.1x & $33.92 \pm 0.06$ \\
\bottomrule
\end{tabular}
\caption{Results for IWSLT'14 sparse DC-Transformer training. All settings use a microbatch size of 250 tokens and \textsc{checkpoint-residual-1}. Memory reduction is computed relative to the dense baseline with 32-bit precision, a minibatch size of 4,000 tokens, no microbatching, and no checkpointing, which uses 2,896 MB [\Cref{fig:pie}(b)] and attains a BLEU score of 35.15. (Note: Nonzero \% refers to the percentage of nonzeros in the fully-connected weight matrices and embeddings.)}
\label{table:transformer}
\end{center}
\end{table}
\aboverulesep=0.4ex
\belowrulesep=0.65ex
\renewcommand{\arraystretch}{1.0}

As is apparent from \Cref{fig:transformer} and \Cref{table:transformer}, for each level of sparsity, training with FP16 arithmetic causes a minor BLEU score drop compared to training with FP32; this gap increases as the percentage of nonzeros reduces. However, from the perspective of \textit{memory usage}, using FP16 is still preferable to using FP32, as it reduces the memory usage by nearly 2x at the cost of a small drop in model quality; as illustrated in Figure \ref{fig:transformer}(b), for a fixed amount of memory, FP16 training achieves a significantly higher BLEU score than FP32.

\subsubsection{Microbatching with Low Precision}
\label{sec:transformer-micro2}
As mentioned above, because the training procedure is mathematically identical whether or not microbatching is used, for computational reasons we actually run all experiments with the standard minibatch size of 4,000 tokens as usual, even though we calculate the memory usage based on a microbatch size of 250 tokens. In other words, our ``microbatching simulation" in this case is the same as standard training. To justify this, we validate that the numerical differences introduced by microbatching and gradient accumulation in FP16 do not affect the BLEU score. Similarly to \Cref{sec:resnet-micro2}, we compare to ``true microbatching" experiments in which FP16 is used and microbatches of size 250 are actually run sequentially through the network and their gradients accumulated. We find that this has no effect on the final model quality: the BLEU score is $\mathbf{35.06 \pm 0.16}$ after training with FP16 and true microbatching, versus $\mathbf{34.99 \pm 0.19}$ after standard training with FP16.

\subsubsection{Baselines}
\begin{figure}[ht]
\vskip 0.2in
\begin{center}
\centerline{\includegraphics[scale=0.5]{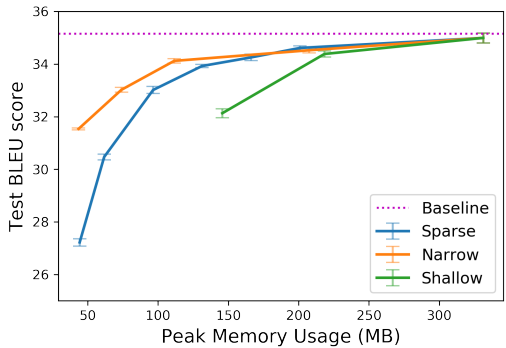}}
\caption{Plot of test BLEU score vs. training memory usage for DC-Transformer on IWSLT'14. The three curves on each plot
are generated by varying the number of layers (``shallow"), the width of the layers (``narrow"), or the sparsity level of the layers (``sparse").
All settings use FP16, a microbatch size of 250 tokens, and \textsc{checkpoint-residual-1}.}
\label{fig:transformer-compare}
\end{center}
\vskip -0.2in
\end{figure}

In Figure \ref{fig:transformer-compare}(a), we compare sparse DC-Transformers to the baseline of smaller DC-Transformer models, reducing either width or depth. (In this case, the width is the dimension of the fully-connected layers in the model, which is equal to the embedding output dimension.) As done for sparsity levels, for each architecture setting we search the base learning rate in \{0.5, 1, 2, 4, 8\} times the default. All models in this comparison are trained in FP16.

For DC-Transformer, we find that, in general, simply making the base model narrower actually performs better than sparsity, in terms of the BLEU score achieved for a fixed memory budget. On the other hand, making the architecture shallower but not narrower, i.e. reducing the number of residual layers, can only reduce memory up to a certain point, because the \textit{embedding} layers take up a significant portion of the memory and these layers cannot be removed. In terms of the memory-BLEU tradeoff, making the model shallower performs slightly worse than sparsity.

\subsubsection{Takeaways}
In summary, we find that sparsity is less effective overall than simply reducing the layer sizes of this DC-Transformer model, as sparsity causes a slightly larger drop in model quality as measured by the BLEU score. Fortunately, FP16, microbatching, and, of course, checkpointing, are very effective, incurring little quality decrease. These three techniques alone give us a 8.7x reduction in memory compared to the baseline, with a BLEU score drop of only 0.15.

\section{Related Work}
Our work extends and synthesizes previous work from four main areas: training sparse neural networks, low-precision training, small-batch training, and gradient checkpointing. We also briefly discuss other techniques for neural network memory reduction.

\subsection{Sparse Neural Networks}
Model sparsity has been studied extensively in the context of compressing trained neural networks. \citet{deepcompression} studied pruning (and quantizing) trained neural networks, iteratively retraining after each pruning and quantization operation. \citet{pruning} improved upon this scheme by incorporating sparsification as part of training, gradually increasing the sparsity level over time by pruning after every few steps of training. \citet{stateofsparsity} compared more advanced sparsity-inducing techniques ($l_0$ regularization and variational dropout) to this
simpler pruning-based scheme on modern large-scale image classification and machine translation tasks, and found that the latter, simpler
scheme generally outperforms the former two. However, all of these works focused on techniques that reduce the memory requirements during \textit{inference}; the methods to \textit{train} these sparse neural networks involve training an initially \textit{dense} model and increasingly sparsifying it, and therefore require the same amount of memory as the standard training approach.

To address this, recent techniques have been proposed to train sparse neural networks under a fixed parameter budget. Perhaps the simplest way to do this is to fix an arbitrarily chosen sparsity pattern \textit{a priori} that does not change during the entire course of training. However, this has been shown to result in a significant decrease in accuracy \citep{lottery}. Recently, \citet{set} and \citet{deepr} independently proposed methods to train sparse neural networks with a fixed number of parameters by adaptively evolving the sparsity pattern. \citet{dsr} (which, with minor modifications, is the sparse training technique we study) extended this work to modern models such as ResNets; the resulting sparse models had competitive accuracy with the corresponding dense model trained normally at sparsity levels up to 90\%. Another approach to training models with a fixed parameter budget was proposed in \citep{golub2019dropback}; in this approach, as with the others, only a fraction of parameters are stored, while the rest are dynamically regenerated to their initial values during passes through the network rather than being set to zero. However, all of these works only studied model sparsity, which is different from our setting of training with a fixed overall memory budget. As illustrated in Figure \ref{fig:pie}, the activation memory can often far outstrip the memory needed even for a dense, uncompressed model, so imposing model sparsity alone does not yield an overall low-memory training method.

\subsection{Low Precision Training}
Another line of work aims to train very low-precision -- even binary (1-bit) -- neural networks from scratch, such as in \citep{qnn, bnn, ternary, q2}. However, these works maintain a full-precision copy of the weights during training, which is necessary in order to accumulate the gradient updates without catastrophic loss of information. As a result, the model memory storage requirements during training are not reduced. Moreover, the most aggressive (lowest-precision) levels of quantization also cause a fairly significant loss in accuracy compared to the state-of-the-art.

Training with half precision has also been extensively studied. \citet{mixedprecision} demonstrated how to train to full accuracy with most storage in half precision (16 bits); we adopt these techniques in this work. However, this work maintains a 32-bit copy of the weights, which actually \textit{increases} the storage required for the module. This was not necessary in the settings we experimented on, as shown in Sections \ref{sec:approach} and \ref{sec:experiments}.

Even if all storage is done in half precision, the memory savings can be at most 2x - an order of magnitude less than the compression ratios achievable by integrating sparsity and batch size reduction as well. Recently, \citet{8bit} showed that all computation during training, aside from accumulation in certain intermediate operations, can actually be done in 8-bit precision; however, they still maintain a 16-bit ``master" copy of the weights.

\subsection{Small-batch Training}
\citet{masters2018revisiting} studied varying the minibatch size for training of neural networks for image classification and found that using minibatches of sizes 2-8 typically yielded the best test set performance. \citet{martinmahoney} corroborated this phenomenon and argued that small-minbatch training benefits generalization performance due to regularization effects. However, changing the batch size usually also requires modifying the learning rate and sometimes other hyperparameters. Moreover, it is unclear whether the generalization benefits of small-minibatch training also apply when sparsity and/or low precision are used, or extend to different settings such as machine translation, although exploring these questions is an interesting direction for future work. Thus, microbatching is a common alternative. The idea of splitting minibatches into smaller microbatches for which gradients are computed independently and accumulated has been proposed several times, although the term \textit{microbatch} originates from \citep{huang2018gpipe}. Indeed, this is how most deep learning frameworks handle multi-GPU execution. \citet{lym2019serialization} proposed a technique to reduce memory traffic by microbatching, via more sophisticated data reuse mechanisms. \citet{ghostbatch} studied the effects of microbatching on neural networks for image classification containing batch normalization and found that it improved generalization performance, although the microbatch sizes studied were themselves comparatively large (128 or 256).

Several alternatives to batch normalization have been proposed that are independent of the microbatch size or robust to even smaller microbatch sizes -- down to 1 or 2 -- such as streaming normalization \citep{streamnorm}, batch renormalization \citep{batchrenorm}, group normalization \citep{groupnorm}, activation normalization \citep{glow}, switchable normalization \citep{switchablenorm}, and fixup initialization \citep{fixupinit}. In the image classification setting, most of these techniques still exhibit a loss of accuracy when used with very small microbatch sizes, compared to the baseline architecture trained with batch normalization and without microbatching. Nevertheless, it is another interesting area for future work to explore the utilization of these approaches to further bolster the accuracy of low-memory image classification training.

\subsection{Gradient Checkpointing}
Gradient checkpointing was originally introduced by \citet{revolve}, and studied in the context of deep learning by \citet{chen2016training}. It is commonly used in practice due to the availability of implementations on modern frameworks such as TensorFlow and PyTorch, and due to the fact that it reduces memory usage without any effects on model accuracy. \citep{openai-blog} suggested several practical checkpointing strategies for architectures such as residual networks, which we adopt in this work. \citet{inplacebn} suggested an improved checkpointing strategy for batch normalization layers, and furthermore proposed a fused normalization and activation layer to entirely obviate the need for storing up to 50\% of the activations.

\citet{gomez2017reversible} introduced reversibility to obviate the need for even storing most network activations, instead allowing
them to be recomputed directly from the next layer's activations in the backward pass;
this technique was adapted and extended by \citet{irevnet}. However, these techniques require modifying model architectures to replace standard network blocks with reversible ones.

\subsection{Other Techniques}
The problem of compressing fully-trained networks has been well-studied. A popular technique is
\textit{quantization}, where weights with similar values are grouped into buckets and assigned to the same
floating-point value so that they can be stored with fewer bits.
Integrated pruning and quantization-based compression techniques are also quite popular and effective
in terms of both compression ratio and preservation of final accuracy compared to the
corresponding uncompressed network \citep{deepcompression}.
In these techniques, small weights are pruned to induce sparsity, and then the remaining
the weights of a trained network are quantized down to a small number of bits with $k$-means quantization.
After fine-tuning, the test accuracy of the resulting compressed network is on par with the original.
More recent techniques based on similar principles include \citep{structured, udit, haq, coreset}.
However, all these techniques require training a full-size network at first,
which is compressed afterwards, so the memory requirements of training remain unchanged.

Similarly, \textit{model distillation}, or knowledge distillation, is a technique for ``compressing" a larger model into a smaller one
by training a smaller model to imitate the larger one; it has been shown to attain better performance than
training from scratch in a variety of settings \citep{distillation}. Again, this technique is only applicable to reducing
inference memory, as it requires first training the larger model.

Others have explored the design of architectures that are dense yet parameter-efficient,
able to attain similar performance (albeit typically with some accuracy loss) to much larger models
when trained from scratch. These models are often designed with both memory and inference speed in mind;
examples include SqueezeNet \citep{squeezenet}, MobileNet \citep{mobilenet, mobilenetv2},
and MorphNet \citep{morphnet}.
\citet{squeezenet} also showed that post-training model compression
could be applied to compress SqueezeNet even further without loss of accuracy.

Another way to reduce the model (and optimizer) memory required for storing and training a neural network
is to replace weight matrices with special \textit{structured} matrices, such as
low-rank matrices \citep{lowrank}, Toeplitz-like matrices \citep{sindhwani}, block-circulant matrices \citep{circulant, circnn},
\textit{Fastfood} transforms \citep{fried}, low displacement rank matrices \citep{ldr},
and butterfly matrices \citep{butterfly}.

To reduce \textit{optimizer} memory specifically, space-efficient alternatives to Adam and other adaptive optimizers
have recently been proposed, such as Adafactor \citep{adafactor} and SM3 \citep{sm3}.
In these approaches, instead of storing momentum values for each weight value,
the parameters are divided into groups and one momentum value is stored for each group.

Another way to reduce \textit{activation} memory, rather than just reducing the number of activations to store,
is to quantize the activations during the forward pass. This has been studied by, for instance, \citet{zipml, gist, gural, q2}.
Similarly, the activations can be sparsified, as in \citep{compressdma, dsg}.
Other works have explored different memory allocation and offloading polices to reduce the activation memory
footprint of training \citep{gist, chen2019modnn, buddy}.
It is an interesting area for future work to apply such activation compression methods
alongside our other techniques, as these methods are orthogonal to the techniques we study in this work.

\section{Conclusion}
In this paper, we study various techniques to reduce the memory requirements of training neural network models. We focus on reducing the model memory, optimizer memory, and activation memory, and investigate the tradeoffs incurred by the explored techniques in terms of the total memory required for training and the final accuracy achieved. We show how to combine these techniques appropriately in order to achieve near-state-of-the-art results, using comparable computation but an order of magnitude less memory than standard training approaches.

\section*{Acknowledgements}
We gratefully acknowledge the support of DARPA under Nos. FA87501720095 (D3M) and FA86501827865 (SDH), NIH under No. U54EB020405 (Mobilize), NSF under Nos. CCF1763315 (Beyond Sparsity) and CCF1563078 (Volume to Velocity), ONR under No. N000141712266 (Unifying Weak Supervision), the Moore Foundation, NXP, Xilinx, LETI-CEA, Intel, Google, NEC, Toshiba, TSMC, ARM, Hitachi, BASF, Accenture, Ericsson, Qualcomm, Analog Devices, the Okawa Foundation, and American Family Insurance, and members of the Stanford DAWN project: Intel, Microsoft, Teradata, Facebook, Google, Ant Financial, NEC, SAP, VMWare, and Infosys. The U.S. Government is authorized to reproduce and distribute reprints for Governmental purposes notwithstanding any copyright notation thereon. Any opinions, findings, and conclusions or recommendations expressed in this material are those of the authors and do not necessarily reflect the views, policies, or endorsements, either expressed or implied, of DARPA, NIH, ONR, or the U.S. Government.

\newpage
\nocite{hashednet}
\nocite{scnn}
\bibliographystyle{plainnat}
\bibliography{low_memory_training}

\newpage
\def\thesection{\Alph{section}}

\appendix
\section*{Appendix}
\crefalias{section}{appsec}

\section{Experimental Details}
\subsection{Experimental Statistics}
\label{app:stats}
For each WideResNet experiment, we run 5 independent trials with different random seeds to obtain the reported results for each hyperparameter setting. For each DC-Transformer experiment, we run 3 independent trials with different random seeds, as the variability between runs is less on DC-Transformer than it is on WideResNet. 95\% confidence interval half-width is computed as the standard error times 1.96.

\subsection{Hyperparameters}
\label{app:hyperparams}
Default hyperparameters for (a) CIFAR-10 experiments [taken from \citep{dsr}] and (b) IWSLT'14 experiments [taken from \citep{payless}] are listed in Table \ref{tb:hyperparams}.\footnote{Original implementation of \citep{dsr} can be found at \url{https://gitlab.com/anon-dynamic-reparam/iclr2019-dynamic-reparam/}. Original implementation of \citep{payless} can be found at \url{https://github.com/pytorch/fairseq/}.} (Any other hyperparameters not explicitly listed in this paper are also kept the same as in the corresponding cited works.)

\begin{table*}[t]
\caption{Hyperparameters for all experiments presented in the paper.}
\label{tb:hyperparams}
\centering
\centering
\setlength\tabcolsep{2.5pt}
\begin{tabular}{l | r r | r r r r }
  \toprule
    Experiment 
    & \multicolumn{2}{c|}{ 
      \begin{tabular}[t]{@{}c}WRN-28-2 \\ on CIFAR-10\end{tabular}
    }
    & \multicolumn{2}{c}{ 
    \begin{tabular}[t]{@{}c} DC-Transformer \\ on IWSLT'14\end{tabular}
    } \\ \midrule \midrule
  
  \multicolumn{7}{c} {Hyperparameters for training} \\ \midrule 
    \begin{tabular}[t]{@{}l}Total number of training updates \end{tabular} 
    & \multicolumn{2}{r|}{100,000 (200 epochs)}  
    & \multicolumn{2}{r}{50,000}  
    \\ \midrule
  \begin{tabular}[t]{@{}l} Base learning rate schedule \\ (epoch/update range: base learning rate)\end{tabular} 
      & \begin{tabular}[t]{@{}r@{}@{}@{}}
        Epochs 1 - 60: \\
        61 - 120: \\ 
        121 - 160: \\
        161 - 200: \\
      \end{tabular}  
    & \begin{tabular}[t]{@{}r@{}@{}@{}}
        0.100 \\ 
        0.020 \\ 
        0.040 \\
        0.008 \\
     \end{tabular} 
     & \begin{tabular}[t]{@{}r@{}@{}@{}}
      \end{tabular} 
    & \hspace{-7em}\begin{tabular}[t]{@{}r@{}@{}@{}}
        Updates $n = 1-4000$:\\ 
        \quad \small{$10^{-7} + \frac{(5*10^{-4} -10^{-7})n}{4000}$} \\ 
        \\
        Updates $n > 4000$: \\ 
        \quad \small{$5*10^{-4} * \sqrt{\frac{4000}{n}}$} \\
      \end{tabular} 
    \\  \midrule
  Optimizer
    & \begin{tabular}[t]{@{}r@{}@{}@{}}SGD; Nesterov \\ momentum 0.9 \end{tabular} &
     \begin{tabular}[t]{@{}r@{}@{}@{}}\end{tabular} &  \begin{tabular}[t]{@{}r@{}@{}@{}}\end{tabular}
    & \hspace{-3em}\begin{tabular}[t]{@{}r@{}@{}@{}}Adam; \\ $\beta_1 = 0.9, \beta_2 = 0.98$ \end{tabular}
    \\ \midrule  
  Weight decay multiplier 
    & \multicolumn{2}{r|}{0.0005}  
    & \multicolumn{2}{r}{0.0001}
    \\ \midrule \midrule
  \multicolumn{7}{c} {Hyperparameters for dynamic sparse reparameterization (DSR)} \\ \midrule
  Default desired fraction of parameters to prune  
    & \multicolumn{2}{r|}{0.01377866}  
    & \multicolumn{2}{r}{0.01377866}
    & \multicolumn{2}{r}{} 
    \\ \midrule  
  Initial pruning threshold 
    & \multicolumn{2}{r|}{0.001}  
    & \multicolumn{2}{r}{0.001}  
    \\ \midrule  
  Pruning threshold adjustment scaling factor
    & \multicolumn{2}{r|}{2}
    & \multicolumn{2}{r}{2}
    \\ \midrule  
  \begin{tabular}[t]{@{}l} Base rewiring schedule \\ (epoch/update range: rewire every X updates) \end{tabular} 
    & \begin{tabular}[t]{@{}r@{}@{}@{}}
        Epochs 1 - 25: \\ 
        26 - 80: \\
        81 - 140: \\
        141 - 190: \\
        190 - 200:
      \end{tabular} 
    & \begin{tabular}[t]{@{}r@{}@{}@{}}
        100 \\ 
        200 \\ 
        400 \\
        800 \\
        0
      \end{tabular}  
    & \begin{tabular}[t]{@{}r@{}@{}@{}}
        Updates 0 - 6250:\\ 
        6250 - 20000:\\ 
        20000 - 35000:\\
        35000 - 47500:\\
        47500 - 50000:\\
      \end{tabular}  
    & \begin{tabular}[t]{@{}r@{}@{}@{}}
        100 \\ 
        200 \\ 
        400 \\
        800 \\
        0 \\
      \end{tabular}  
  \\ \bottomrule

\end{tabular}  
\end{table*}

\subsection{Hyperparameter Tuning Procedure}
\label{app:validation}
The hyperparameters we tune for different network architectures and sparsity levels are the learning rate and, in the case of sparse networks, the rewiring frequency and rewiring fraction. We tune these for each architectural setting, i.e. each network with a different sparsity level, number of layers, or layer width. We apply the chosen parameters across all experiments with the same architectural setting.

For WideResNet CIFAR-10 hyperparameter searches, we select the best hyperparameters by training on 45,000 of the 50,000 training examples and validating on a held-out validation set randomly sampled on the training data. We run 5 independent trials with different random seeds for each setting, and pick the hyperparameter or combination of hyperparameters yielding the best mean validation accuracy. For the test set experiments, we then train on the full set of 50,000 training examples and test on the predefined held-out test set of 10,000 examples.

For each WideResNet network architecture or sparsity level, we search the learning rate in \{0.25, 0.5, 1, 2\} times the default from \citep{dsr} (see \Cref{tb:hyperparams}). Results are presented in \Cref{table:cifar-lr}. Details on how we tune the rewiring frequency and rewiring fraction are provided in \Cref{sec:dsr-details}.

For DC-Transformer IWSLT'14 hyperparameter searches, we select the best hyperparameters by training on the predefined training set and validating on the separate, predefined validation set, as done in \citep{payless} and the \texttt{fairseq} repository. We run 3 independent trials with different random seeds for each setting, and pick the hyperparameter or combination of hyperparameters yielding the best mean validation BLEU score, as we found 3 trials to be sufficient to determine the best hyperparameter within 95\% confidence in this setting. For the test set experiments, we test on the predefined held-out IWSLT'14 test set, again as done in \citep{payless} and the \texttt{fairseq} repository.

For each DC-Transformer network architecture or sparsity level, we search the learning rate in \{0.5, 1, 2, 4, 8, 16\} times the default from \citep{payless} (see \Cref{tb:hyperparams}). Results are presented in \Cref{table:iwslt-lr}. Details on how we tune the rewiring frequency and rewiring fraction are provided in \Cref{sec:dsr-details}.

\subsubsection{Dynamic Sparse Reparameterization Details}
\label{sec:dsr-details}
\paragraph{WideResNet (CIFAR-10)}

The default DSR hyperparameters are the same as in \citep{dsr}. For the WideResNet on CIFAR-10, \citet{dsr} set the desired number of parameters to prune at each step to 20,000. In our reimplementation, we instead specify the amount of parameters to rewire as a \textit{fraction} of the overall number of elements in the tensors to be sparsified, rather than a raw number. This has the benefit of being directly generalizable to neural networks with different numbers of parameters (such as DC-Transformer). We set this fraction to 0.01377866 by default because this corresponds to 20,000 parameters on the CIFAR-10 WideResNet architecture, and therefore matches \citep{dsr}. As mentioned in \Cref{sec:sparsity}, for each sparsity level, we then grid search for the best rewiring frequency and rewiring fraction, where the grid points are \{0.125, 0.25, 0.5, 1, 2, 4\} times the default frequency from \citep{dsr} and \{0.25, 0.5, 1, 2, 4, 8\} times the default rewiring fraction from \citep{dsr}. [Multiplying the rewiring frequency by $k$ means that, for each period in the rewiring schedule, instead of the default which is rewiring every $X$ updates, we rewire every $X / k$ updates; so, a larger rewire frequency multiplier means rewiring is done more often.] For each sparsity level, we select the best combination of these hyperparameters, as determined by validation accuracy, and use it for all test set experiments with that sparsity level. The chosen hyperparameters for each sparsity level are presented in \Cref{table:cifar-dsr}.

\paragraph{DC-Transformer (IWSLT'14)}
For DC-Transformer, we set the default DSR hyperparameters to be the same as in \citep{dsr}, and thus the same as those used for the WideResNet. The default rewiring schedule, however, must be adapted due to the fact that the standard training schedule for DC-Transformer is for half as many updates (50,000) as the number of updates we train the WideResNet for. To address this, we simply halve the length of each period in the rewiring schedule; for instance, the first rewiring period for training WideResNet is from epochs 1-25 (updates 0-12500), so training DC-Transformer we make the first rewiring period to be from updates 0-6250. As done for WideResNet, for each sparsity level, we conduct a grid search over the rewiring frequency and rewiring fraction, where the grid points are \{0.0625, 0.125, 0.25, 0.5, 1, 2\} times the default frequency from \citep{dsr} and \{0.25, 0.5, 1, 2, 4, 8\} times the default rewiring fraction from \citep{dsr}; we additionally evaluated the option of not rewiring at all (and found it to give suboptimal BLEU score results for all sparsity levels). For each sparsity level, we select the best combination of these hyperparameters, as determined by validation BLEU score, and use it for all test set experiments with that sparsity level. The chosen hyperparameters for each sparsity level are presented in \Cref{table:iwslt-dsr}.

\begin{table}
  \small
  \begin{center}
    \setlength{\tabcolsep}{20pt}
    \begin{tabular}{@{}rrrr@{}}
      \toprule
       \% Nonzero & \# Layers & Width Multiplier & Learning Rate Multiplier  \\
      \midrule
      100	&28		&2		& 0.5   \\
      50	&28		&2		& 0.5  \\
      40	&28		&2		& 1   \\
      30	&28		&2		& 1  \\
      20	&28		&2		& 1   \\
      10	&28		&2		& 0.5  \\
      5 	&28		&2		&0.5  \\
      100	&22		&2		&0.5  \\
      100	&16		&2		&0.5  \\
      100	&10		&2		&0.5  \\
      100	&28		&1.5		&0.5  \\
      100	&28		&1		&0.5  \\
      100	&28		&0.75	&0.5  \\
      100	&28		&0.5		& 1  \\
      \bottomrule
    \end{tabular}
  \end{center}
      \caption{Learning rate multipliers used for WideResNet experiments. Recall from \Cref{tb:hyperparams} that the initial
      learning rate is 0.1, with step decays after epochs 60, 120, and 160. For each architecture or sparsity setting,
      we multiply this learning rate by the multiplier in the above table.}
      \label{table:cifar-lr}
\end{table}

\begin{table}
  \small
  \begin{center}
    \setlength{\tabcolsep}{20pt}
    \begin{tabular}{@{}rrrr@{}}
      \toprule
       \% Nonzero & \# Decoder Layers & Embedding Dimension & Learning Rate Multiplier  \\
      \midrule
      100	&6		&512		&1   \\
      50	&6		&512		&2  \\
      40	&6		&512		&2   \\
      30	&6		&512		&4  \\
      20	&6		&512		&4   \\
      10	&6		&512		&8  \\
      5 	&6		&512		&8  \\
      100	&3		&512		&1  \\
      100	&1		&512		&2  \\
      100	&6		&384		&1  \\
      100	&6		&256		&2  \\
      100	&6		&192		&2  \\
      100	&6		&128		&4  \\
      \bottomrule
    \end{tabular}
  \end{center}
      \caption{Learning rates used for DC-Transformer experiments. Refer to \Cref{tb:hyperparams} for the
      base learning rate schedule. For each architecture or sparsity setting,
      we multiply this learning rate by the multiplier in the above table.}
      \label{table:iwslt-lr}
\end{table}

\begin{table}
  \small
  \vskip -0.2in
  \begin{center}
    \begin{tabular}{@{}rrrr@{}}
      \toprule
       \% Nonzero & Rewire Fraction Multiplier & Rewire Frequency Multiplier  \\
      \midrule
      50	&4		&0.5	 \\
      40	&1		&2	   \\
      30	&2		&0.5	  \\
      20	&2		&1	   \\
      10	&1		&0.5	  \\
      5 	&1		&0.25	  \\
      \bottomrule
    \end{tabular}
  \end{center}
      \caption{Dynamic sparse reparameterization hyperparameters used for WideResNet experiments.
      Refer to \Cref{tb:hyperparams} for the default rewiring fraction and frequency; for each sparsity level,
      we multiply these defaults by the multipliers in the above table.}
      \label{table:cifar-dsr}
\end{table}

\begin{table}
  \small
  \vskip -0.05in
  \begin{center}
    \begin{tabular}{@{}rrrr@{}}
      \toprule
       \% Nonzero & Rewire Fraction Multiplier & Rewire Frequency Multiplier  \\
      \midrule
      50	&4		&0.125	 \\
      40	&4		&0.125	   \\
      30	&4		&0.125	  \\
      20	&2		&0.25	   \\
      10	&1		&0.5	  \\
      5 	&0.5		&0.5	  \\
    \bottomrule
    \end{tabular}
  \end{center}
      \caption{Dynamic sparse reparameterization hyperparameters used for DC-Transformer experiments.
      Refer to \Cref{tb:hyperparams} for the default rewiring fraction and frequency; for each sparsity level,
      we multiply these defaults by the multipliers in the above table.}
      \label{table:iwslt-dsr}
\end{table}

\section{Activation Memory Profiler Implementation}
\label{app:profiler-impl}
We built a profiler to calculate the total activation memory and FLOPs required for a single forward-backward pass. The profiler traverses the autograd graph in reverse topological order. Each leaf node in the reversed graph represents input data or a model weight, while each non-leaf node represents a mathematical operation. To compute backward through a node, i.e. compute the gradient of the loss with respect to its input, in general we need the node's inputs as well as the gradients of the loss with respect to the node's outputs. When no checkpointing is used, we simply count up all activations up to each node that must be stored in the forward pass for this procedure to be completed, as well as the memory for the backward activations that must be stored while backpropagating through the node, sum these, and find the maximum result over all such nodes in the graph. As described in \Cref{sec:background}, we filter out nodes that simply reshape or transpose the data, or otherwise do not require their inputs to compute gradients (for instance, addition operations), and handle special cases such as ReLU for which a ``compressed" version of the input can be stored \citep{gist}. Note that we only count the memory for gradients with respect to the forward activations as ``activation memory," as gradients with respect to the \textit{weights} are materialized in the model gradient buffer, which is separately accounted for. We also do not calculate temporary memory for intermediate computations, since this is a small fraction of the total memory usage in general and can be highly dependent on the implementation of these computations.

When checkpointing is used, we count the activation storage for the ``checkpoint nodes" (nodes whose output activations are stored) rather than all nodes in the graph. In addition, we count up the memory required for storing all necessary forward activations in each given ``segment" between consecutive checkpoint nodes. The memory for backward activations is computed as before; again, we compute the maximum amount of memory that must be stored at any one time with the given checkpointing strategy. When counting the number of FLOPs for recomputation, we note that the output of some layers (such as batch normalization layers) can be computed more cheaply than during the original forward pass by caching small quantities \citep{inplacebn}; we account for this additional memory as well (although it turns out to be less than 1\% of the total memory usage).

This strategy in which activations are recomputed as necessary during backpropagation is currently the standard way to implement checkpointing \citep{openai-blog} and is well-suited for the GPU machine model. In future work, we also plan to consider more sophisticated methods that utilize paging to off-device memory in addition to recomputation, which may be better suited for different computing substrates.

\end{document}